\documentclass{article}
\usepackage{spconf,amsmath,graphicx}
\usepackage{xcolor}
\usepackage{adjustbox}
\usepackage{colortbl}
\usepackage{multirow}
\usepackage{makecell}
\usepackage{booktabs}

\title{VT-CLIP: Enhancing Vision-Language Models with Visual-guided Texts}
\name{Longtian Qiu$^{\star}$, Renrui Zhang$^{\dagger}$, Ziyu Guo$^{\dagger}$, Ziyao Zeng$^{\star}$, Zilu Guo$^{\diamondsuit}$, Yafeng Li$^{\nabla}$, Guangnan Zhang$^{\nabla}$}

\address{$^{\star}$ShanghaiTech University,\qquad $^{\dagger}$Peking University,\qquad $^{\diamondsuit}$Nankai University,\\
$^{\nabla}$Baoji University of Arts and Science\\
}

\begin{document}
%
\maketitle
\begin{abstract}
Contrastive Language-Image Pre-training (CLIP) has drawn increasing attention recently for its transferable visual representation learning. However, due to the semantic gap within datasets, CLIP's pre-trained image-text alignment becomes sub-optimal on downstream tasks, which severely harms its transferring performance. To better adapt the cross-modality embedding space, we propose to enhance CLIP via Visual-guided Texts, named VT-CLIP. Specifically, we guide textual features of different categories to adaptively explore informative regions on the image and aggregate visual features by attention mechanisms. In this way, the texts become visual-guided, namely, more semantically correlated with downstream images, which greatly benefits the category-wise matching process. In few-shot settings, we evaluate our VT-CLIP on 11 well-known classification datasets to demonstrate its effectiveness.
\end{abstract}

\section{Introduction}
\label{sec:intro}

With the exploring of better network architectures, traditional deep learning has achieved extraordinary performances over a wide range of vision tasks, such as image classification~\cite{he16}, object detection~\cite{detr} and so on~\cite{he2017mask}. Although these methods are expert at specific scenarios, they lack the ability of general vision representation and are hard to transfer to open-set applications. Considering the wide coverage of languages, CLIP~\cite{rad21} proposes to pre-train the visual representation learning contrastively with natural language signals. Supervised by large-scale image-text pairs, CLIP extracts both features of input images and texts by separate encoders and matches the paired ones in the same embedding space. Such pre-trained cross-modality alignment endows CLIP the capability of recognizing new visual concepts on downstream tasks. Specifically, given a new dataset with images to be recognized, one could construct the textual inputs of CLIP by the category names, termed as prompts, and convert the classification task into a matching problem. By this, CLIP is able to conduct zero-shot recognition in open-vocabulary settings.

To further improve the downstream performance of CLIP, existing works introduce different fine-tuning methods under the few-shot settings. Inspired by prompt tuning in natural language processing~\cite{shin20}, Context Optimization (CoOp)~\cite{zhou2022coop} freezes CLIP's pre-trained weights and adopts learnable prompts to learn the best-fitted textual inputs other than the original hand-crafted ones.
From another perspective, CLIP-Adapter~\cite{clip} appends a lightweight adapter module~\cite{houlsby2019parameter} over the image or text encoders and is also fine-tuned with frozen CLIP's weights. 
However, they are constrained by the following limitations.
The learnable prompts in CoOp are set before the large-scale text encoder, which is much time-consuming to back-propagate the training gradients through such 12-layer transformer~\cite{vaswani} for every iteration.
CLIP-Adapter only conducts feature adaption independently for each branch, and lacks cross-modal interactions for image and language. Also, both of their learned parameters are fixed for all images during inference, losing the adaption flexibility for varying visual inputs.

\begin{figure}[t]

\centering
\centerline{\includegraphics[width=7.5cm]{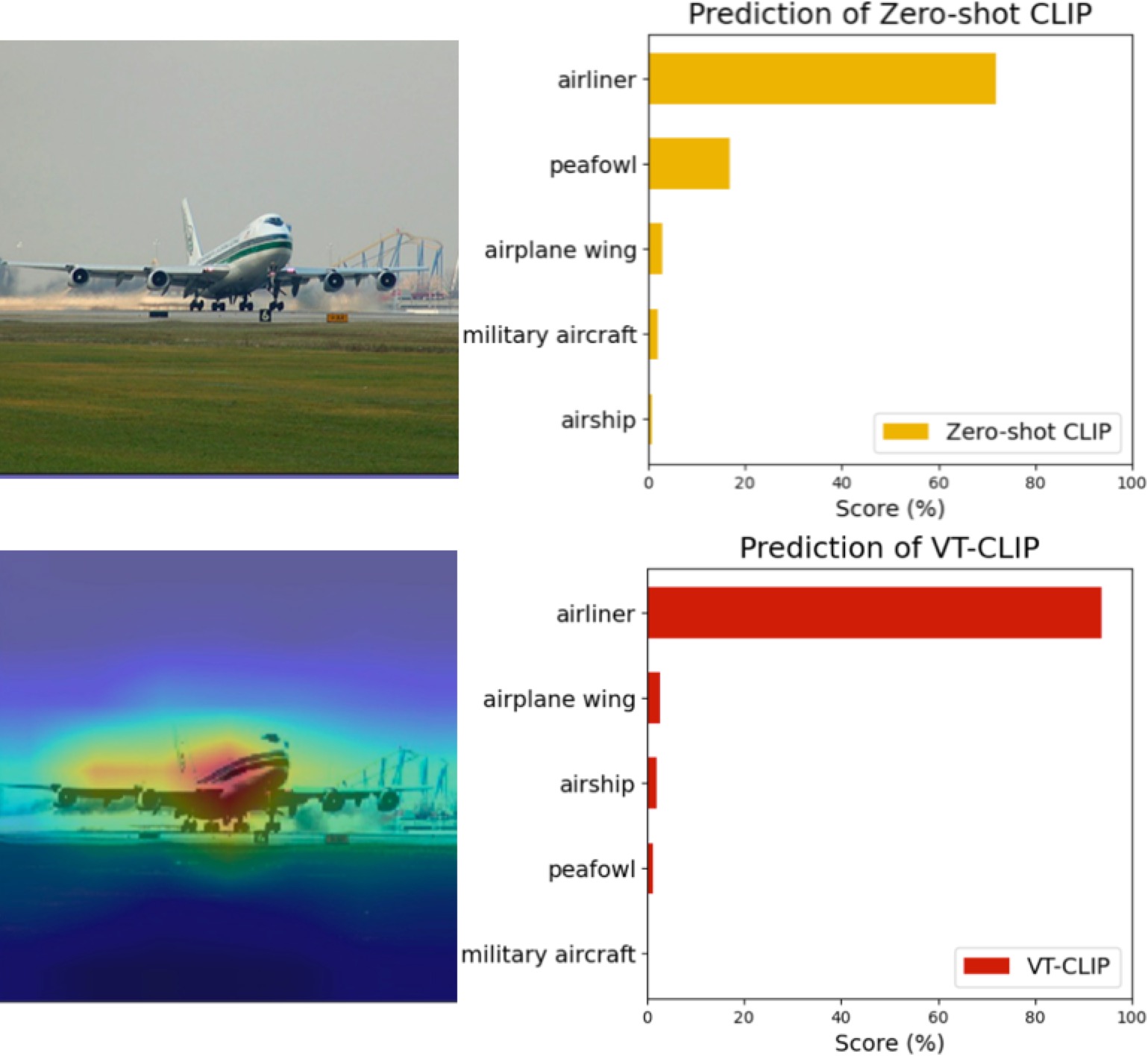}}
\vspace{-0.2cm}
\caption{\textbf{Visualization of attention map and prediction score.} In our VT-CLIP, the informative regions on the image gain more attention weights from the texts, which improves the prediction score on ground-truth category.}

\label{fig:teaser}
\end{figure}

In this paper, we propose \textbf{VT-CLIP}, which equips CLIP with \textbf{V}isual-guided \textbf{T}exts to exert thorough adaptions of both modalities under few-shot settings. Specifically, we introduce a visual-guided attention module to conduct feature communication after both encoders, which views texts as the queries, and images as keys and values. For textual branch, we input the hand-crafted prompts with explicit semantics and utilize the encoded features for interaction with images. For visual branch, we extract the intermediate spatial features of images as keys and values instead of the final global ones, which could provide more fine-grained contextual information. By calculating the per-pixel similarities, different categories in texts could explore informative visual regions and gather related features weighted by their attention scores. After this, the texts are visual-guided and better fitted for the later matching stage. As visualized in Figure~\ref{fig:teaser}, the textual feature of "airline" category focuses on the corresponding visual regions, but other unmatched categories do not, as expected. Importantly, our visual-guided texts are adaptive for different samples, since the attention map is dynamically produced by the input images. We conduct extensive experiments on 11 well-known classification datasets, which fully demonstrates the outstanding enhancement ability of visual-guided texts over CLIP.


\vspace{-0.1cm}
\section{RELATED WORK}

\label{sec:related}
Recently, vision-Language Models shows great potential in learning generic visual representation with nature language supervision, which allowing zero-shot transfer ability for various downstream classification tasks. Inspired by the success of pre-train models~\cite{BERT}, ~\cite{UNITER}~\cite{Oscar} and SimVLM ~\cite{SimVLM} use attention architecture improve the performance of vision-language tasks. At present, the recent breakthrough in vision-language learning, particularly CLIP~\cite{rad21} and ALIGN~\cite{jia21} are driven by the noisy large-scale datasets available in the Internet, which is 400 million image-text pair for CLIP and 1.8 billion noisy image-text pairs for ALIGN.To fine-tune vision-Language Models on downstream tasks like few-shot classification task, CoOp~\cite{zhou2022coop} propose to learn soft prompts represented by continuous context vectors as alternative for hand-craft prompt while CLIP-Adapter propose to adopts an additional bottleneck layer to learn new features and performs residual style feature blending with the original pre-trained features. Though CoOp and CLIP-Adapter achieve significant performance in the perspective of prompt learning and feature adapters, our VT-CLIP explores the impact of instance-level image visual feature on refining text feature with a cross-attention module.

Prompt Learning are designed to better mine the knowledge from pre-trained models without fine-tuning the entire model, which generate a prompting template or function to bridge gap between the pre-training objective and downstream tasks~\cite{shin20,jiang20,liliang,lester}. Prompt engineering is an important topic in prompt learning. Early research focus on designing hand-crafted prompts, which generate cloze-style prompts like “fill-in-the-blank” cloze tests and benefits a number of downstream tasks, such as sentiment analysis~\cite{jiang20}. Recently, ~\cite{liliang,lester} introduce gradient-based approaches which optimize continuous vectors in the word embedding space. The limitation of prompt engineering is that hand-crafted prompt template requires specific domain knowledge and the prompt content learned by optimization lacks interpretability. In this paper, we demonstrate guiding text feature with instance-level image feature through cross-attention module is an alternative for prompt learning on large-scale vision-language models, which is more interpretable and simpler in architecture.

\begin{figure*}[htb]

\centering
\includegraphics[width=1.5\columnwidth]{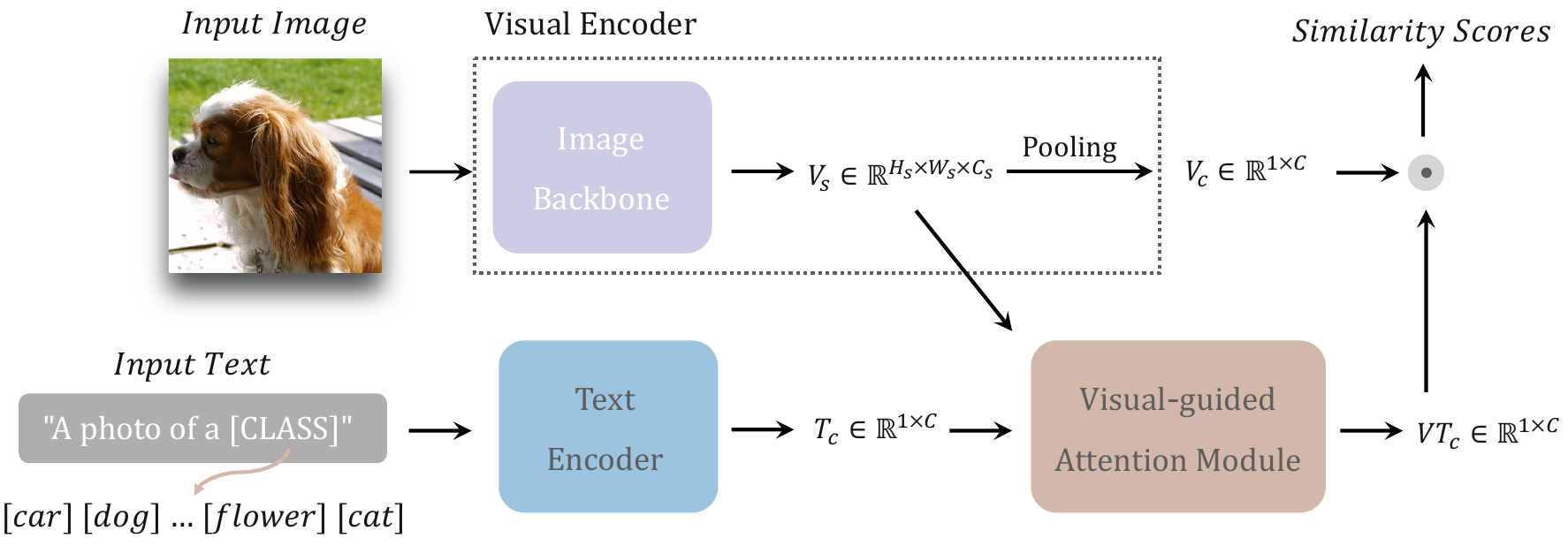}

\caption{\textbf{Structures of VT-CLIP}. Class-level text feature from CLIP text encoder is updated by spatial visual image feature with cross attention module}
\vspace{-0.1cm}
\label{fig:pipeline}
\end{figure*}

\section{METHODS}
\label{sec:method}
 We first revisit the zero-shot CLIP in Section~\ref{s3.1}, and then introduce the details of our proposed VT-CLIP in Section~\ref{s3.2}.
\vspace{-0.15cm}

\subsection {Zero-shot CLIP}
\label{s3.1}
CLIP is pre-trained to align image and text pair information through contrastive training. CLIP contains two independent encoders for visual and textual feature encoding. Specifically, the image encoder consists of a visual backbone, which is ResNet~\cite{he16} or ViT~\cite{dos21}, and an attention pooling layer, while the text encoder is a conventional Transformer Encoder~\cite{vaswani}. With large-scale data traning, informative representation for both modalities are deeply learned by CLIP's image encoder and text encoder, thus obtaining zero-shot classification ability. 
In details, for a dataset contains $K$ categories, denoted as \{$C_1,\dots,C_K$\}, CLIP first places all category names into the hand-crafted template $H$ proposed by~\cite{rad21} to get the textual inputs, which then are fed into the tokenizer $T$ and text encoder to obtain text features, $T_c \in R^{1 \times C}$. Meanwhile, the input image $I \in R^{H\times{W\times{3}}}$, where $H$ and $W$ are the height and width of the image respectively, is first encoded by visual backbone, termed as $VB$, getting contextual-level spatial features $V_s$. After that, an attention pooling operation is adopted to get global visual features $V_c$, i.e.,
\begin{gather}
V_s = \operatorname{VB}(I), V_s\in R^{H_s \times W_s \times C_s} \\
V_c = \operatorname{Pooling}(V_s), V_c\in R^{1 \times C} \\
T_c = \operatorname{TextEn}\big(T([H; C_i])\big), i \in \{1,\dots,K\},
\end{gather}
where $TextEn$ denotes text encoder, and $C$ is the class-level feature dimension. The $H_s, W_s, C_s$ are the height , width and channel dimension for spatial feature.
Via attention pooling, CLIP generates the global visual features from spatial features which contain more local contextual-level information. Finally, the similarity scores are calculated as,
\begin{gather}
P = \operatorname{Softmax}(V_cT_c^T/\tau),
\end{gather}
where $SoftMax(\cdot)$ and $P$ denote the softmax function and the similarity scores for $K$ categories, and $\tau$ is a temperature parameter learned by CLIP.

\subsection {VT-CLIP}
\label{s3.2}
Different from the perspective of prompt learning~\cite{zhou2022coop}, we present a new approach to enhance vision-language model. We suppose that generic soft prompt which is invariant to images with various content is kind of insufficient. Hence, we propose VT-CLIP, which dynamically refine the text features using visual spatial features. Specifically, through a visual-guided cross-attention module, we leverage the contextual-level spatial feature, which is obtained before pooling, to guide the text feature to adaptively explore informative regions on the image. The learned refinement is fused with original text features a by residual connection to preserve the robustness and effectiveness. Following the standard architecture of the transformer decoder blocks \cite{vaswani}, our proposed visual-guided cross-attention module includes a self-attention layer, a co-attention layer and a feed forward network, where $T_c$ and $V_s$ are fed into the cross attention module, with $T_c$ serving as query, and $V_s$ as key and value, i.e.,
\begin{gather}
VT_c = \operatorname{CrossAttn}(V_s, V_s, T_c) + T_c,\\ \notag
VT_c \in R^{1\times C}
\end{gather}
where $CrossAttn$ is the visual-guided cross-attention module, and $VT_c$ denotes the adapted text features. 

Through the interaction in cross-attention, the adapted text features $VT_c$ become more semantically correlated with the paired image. Then, the similarity scores are predicted via obtained $VT_c$, that is,
\begin{gather}
P = \operatorname{Softmax}(V_cVT_c^T/\tau).
\end{gather}

During training, we freeze both visual baskbone and textual encoder, and only optimize weights in visual-guided cross-attention module via
cross-entropy loss.

\begin{figure*}[t]

\centering
\includegraphics[width=1.9\columnwidth]{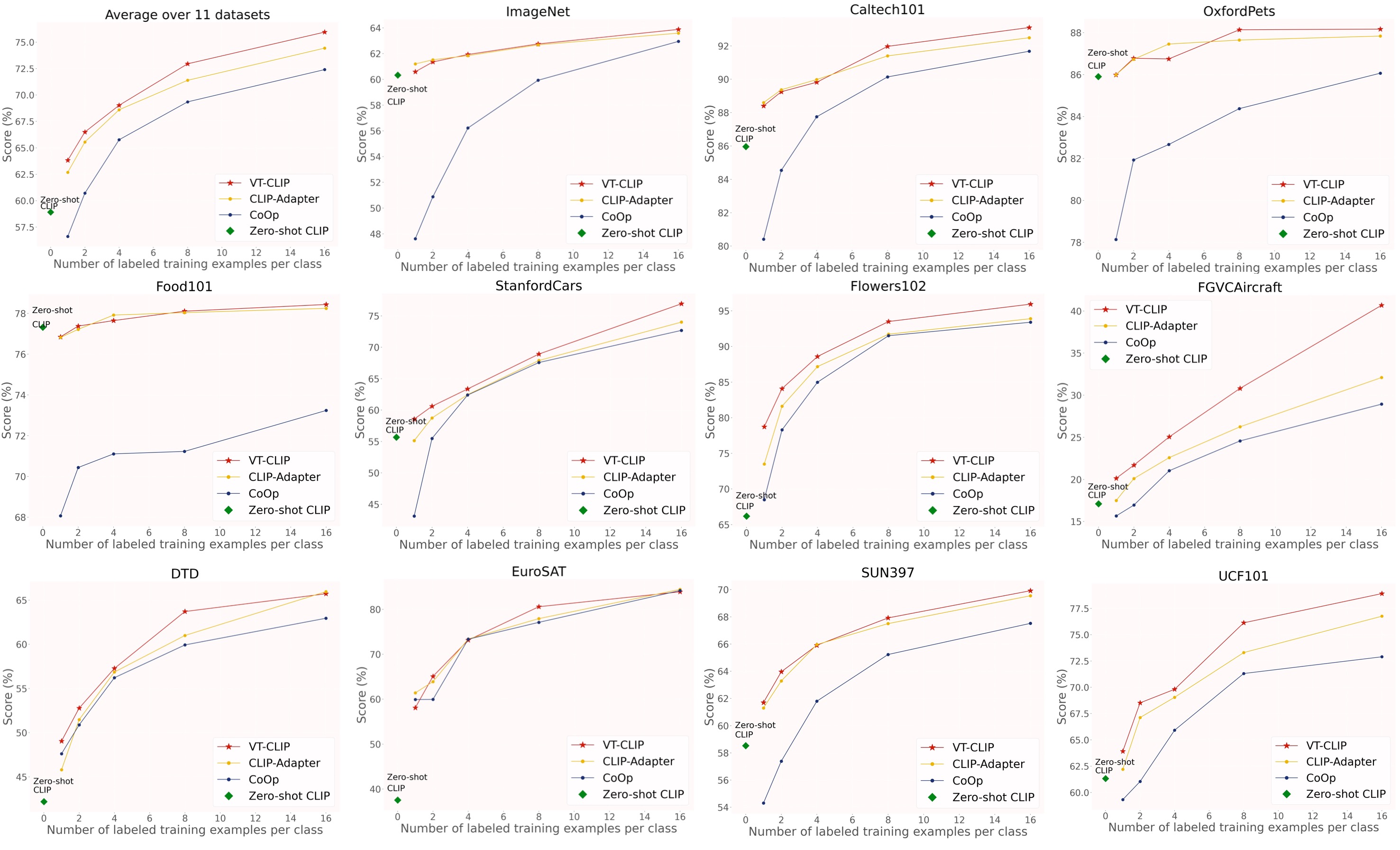}

\caption{\textbf{Experiment Results}––Main results of few-shot learning on 11 datasets. VT-CLIP shows overall better performance over baselines across different training shots.}
\vspace{-0.1cm}
\label{fig:result}
\end{figure*}
\section{EXPERIMENTS}
\label{sec:experiments}
\subsection{Training Settings}
We evaluate the performance of VT-CLIP on 11 widely-adopted image classification datasets and follow the few-shot evaluation protocol of CLIP~\cite{rad21}, that is, training on $1$, $2$, $4$, $8$ and $16$ shots and testing on full test set. 
During training, we adopt pre-trained visual backbone ResNet-50 from \cite{rad21} with all the weights frozen, and follow the data preprocessing protocol of CLIP, which consists of random cropping, resizing, and random horizontal flip. Following CoOp~\cite{zhou2022coop}, VT-CLIP is trained with batch size $32$ and learning rate $2\times10^{-3}$ for all 11 datasets. 
Instead of using learnable continuous prompts in CoOp, we adopt the same hand-crafted prompt as CLIP. We compare VT-CLIP with three baseline works, i.e., Zero-shot CLIP~\cite{rad21}, CoOp~\cite{zhou2022coop} and CLIP-Adapter~\cite{clip}. Also, in order to thoroughly demonstrate the superiority of proposed VT-CLIP, we use the best baseline variants. 

\vspace{-0.3cm}
\subsection{Performance Comparison \& Analysis}
The main results are presented in Figure \ref{fig:result}. As shown in the top-left chart of Figure \ref{fig:result}, VT-CLIP shows outstanding average performance over three baselines, and the accuracy gain increases as the training shots get more, which indicates VT-CLIP serves as an effective and reliable enhancer under few-shot settings.
Also, as shown in all the twelve charts in Figure \ref{fig:result}, our VT-CLIP outperforms other works significantly under each shot setting on 11 datasets. What's more, unlike CoOp's poor performance with little training samples, VT-CLIP achieves more stable scores, which indicates our VT-CLIP is not sensitive to data scale. Additionally, it is clear that VT-CLIP obtains consistently prominent results on all the 11 datasers, which demonstrates more considerable generalization ability than CoOp, as seen in charts of OxfordPets~\cite{parkhi} and Food101~\cite{bossard}, where CoOp falls behind even zero-shot CLIP under 16-shot setting. As for CLIP-Adapter, VT-CLIP not only surpasses it on different datasets, but also contains better interpretability of leveraging contextual visual features to guide text features to be more semantically correlated to the certain downstream task. 
The consistent superiority of VT-CLIP over 11 datasets fully demonstrates the effectiveness and generality of our proposed method.


\section {Ablation Study}
\vspace{-0.15cm}
In this section, we conduct several ablation studies for VT-CLIP. All experiments below adopt the 16-shot setting.


\begin{table}
\centering
\small
\begin{tabular}{lcccc}
\toprule
Heads & 4 & 8 & 16 & 32 \\
\cmidrule(lr){1-1}\cmidrule(lr){2-5}  
Caltech101 (\%) & 93.06 & \textbf{93.10} & 92.37 & 92.62 \\
DTD (\%) & 64.42 & \textbf{65.72} & 64.78 & 65.43 \\
\bottomrule
\end{tabular}

\caption{
\textbf{Head Number.} Performance with different number of heads in cross attention.
} 
\label{tab:heads}
\end{table}

\noindent We explore the number of heads in cross attention on Caltech-\\101~\cite{feifei04} and DTD \cite{cimpoi}. The heads number in attention mechanism equals to the number of heterogeneous scores computed for values indicating fitting ability of model. As presented in Table \ref{tab:heads}, the best performance is achieved with two heads for cross attention module. To further demonstrate the design of our method, we conduct experiments on increase the number of cascaded attention layers. The results are in Table \ref{tab:layers}. We observe that performance degrades as the cascaded layers increase which indicate the complex model with multiple cross attention layer tend to overfit the insufficient training data under few shot scenario.

\begin{table}
\centering
\small
\begin{tabular}{lcccc}
\toprule
Layers & 1 & 2 & 3 & 4 \\
\cmidrule(lr){1-1}\cmidrule(lr){2-5}  
Caltech101(\%) & \textbf{93.10} & 93.06 & 92.58 & 92.29 \\
DTD (\%) & \textbf{65.72} & 64.60 & 65.60 & 64.78 \\
\bottomrule
\end{tabular}

\caption{
\textbf{Layer Number.} Performance with different number of cascaded cross attention layers.
\vspace{0.15cm}
} 
\label{tab:layers}
\end{table}

\section{CONCLUSION}
\vspace{0.15cm}
\label{sec:conclusion}
We propose VT-CLIP, a novel enhancement of CLIP for few-shot classification which focuses on leveraging the contextual visual features to guide the text features to highlight the important regions via a visual-guided cross-attention module. 
In this way, the deep interaction between the image and text branches in vision-language model is of great potential in enhancing the model's ability. Also, extensive experiments demonstrate that VT-CLIP outperforms all the competitive baselines in few-shot settings on 11 widely-used datastes. Ablation studies are conducted to further prove our design and give a view of the extensive performance of VT-CLIP. In the future, we hope to combine VT-CLIP with prompt learning based approaches to push the boundary further for fine-tuning vision-language models. We will also explore the potential of VT-CLIP on other vision or textual tasks.







\vfill\pagebreak


\bibliographystyle{IEEEbib}
\bibliography{icassp}

\end{document}